\documentclass[pdflatex,sn-nature]{sn-jnl}% Style for submissions to Nature Portfolio journals
\usepackage{graphicx}%
\usepackage{multirow}%
\usepackage{amsmath,amssymb,amsfonts}%
\usepackage{amsthm}%
\usepackage{mathrsfs}%
\usepackage[title]{appendix}%
\usepackage{xcolor}%
\usepackage{pifont} % for \xmark and \cmark
\usepackage{textcomp}%
\usepackage{manyfoot}%
\usepackage{booktabs}%
\usepackage{algorithm}%
\usepackage{algorithmicx}%
\usepackage{algpseudocode}%
\usepackage{listings}%
\usepackage[utf8]{inputenc}
\usepackage{textgreek}
\usepackage[table]{xcolor} % add this in the preamble
\usepackage[T1]{fontenc}    % use 8-bit T1 fonts
\usepackage{hyperref}       % hyperlinks
\usepackage{url}            % simple URL typesetting
\usepackage{nicefrac}       % compact symbols for 1/2, etc.
\usepackage{microtype}      % microtypography
\newcommand{\cmark}{\ding{51}}  % ✓
\newcommand{\xmark}{\ding{55}}  % ✗
\theoremstyle{thmstyleone}%
\theoremstyle{thmstyletwo}%

\theoremstyle{thmstylethree}%

\begin{document}

\title[Article Title]{Synergistic Neural Forecasting of Air Pollution with Stochastic Sampling}

%%=============================================================%%
%% GivenName	-> \fnm{Joergen W.}
%% Particle	-> \spfx{van der} -> surname prefix
%% FamilyName	-> \sur{Ploeg}
%% Suffix	-> \sfx{IV}
%% \author*[1,2]{\fnm{Joergen W.} \spfx{van der} \sur{Ploeg} 
%%  \sfx{IV}}\email{iauthor@gmail.com}
%%=============================================================%%

% \author*[1,2]{\fnm{First} \sur{Author}}\email{iauthor@gmail.com}

% \author[2,3]{\fnm{Second} \sur{Author}}\email{iiauthor@gmail.com}
% \equalcont{These authors contributed equally to this work.}

% \author[1,2]{\fnm{Third} \sur{Author}}\email{iiiauthor@gmail.com}
% \equalcont{These authors contributed equally to this work.}

% \affil*[1]{\orgdiv{Department}, \orgname{Organization}, \orgaddress{\street{Street}, \city{City}, \postcode{100190}, \state{State}, \country{Country}}}

% \affil[2]{\orgdiv{Department}, \orgname{Organization}, \orgaddress{\street{Street}, \city{City}, \postcode{10587}, \state{State}, \country{Country}}}

% \affil[3]{\orgdiv{Department}, \orgname{Organization}, \orgaddress{\street{Street}, \city{City}, \postcode{610101}, \state{State}, \country{Country}}}

\author[1]{\fnm{Yohan} \sur{Abeysinghe}}
\email{abeysingheyohan@gmail.com}

\author*[1]{\fnm{Muhammad Akhtar} \sur{Munir}}
\email{akhtar.munir@mbzuai.ac.ae}

\author[1]{\fnm{Sanoojan} \sur{Baliah}}
\email{sanoojan.baliah@mbzuai.ac.ae}

\author[2]{\fnm{Ron} \sur{Sarafian}}
\email{ron.sarafian@weizmann.ac.il}

\author[1,3]{\fnm{Fahad Shahbaz} \sur{Khan}}
\email{fahad.khan@mbzuai.ac.ae}

\author[2]{\fnm{Yinon} \sur{Rudich}}
\email{yinon.rudich@weizmann.ac.il}

\author[1,4]{\fnm{Salman} \sur{Khan}}
\email{salman.khan@mbzuai.ac.ae}

\affil[1]{\orgname{Mohamed bin Zayed University of Artificial Intelligence}, 
\orgaddress{\city{Abu Dhabi}, \country{United Arab Emirates}}}

\affil[2]{\orgname{Weizmann Institute of Science}, 
\orgaddress{\city{Rehovot}, \country{Israel}}}

\affil[3]{ 
\orgname{Linköping University}, 
\orgaddress{\city{Linköping}, \country{Sweden}}}

\affil[4]{\orgname{Australian National University}, 
\orgaddress{\city{Canberra}, \country{Australia}}}

%%==================================%%
%% Sample for unstructured abstract %%
%%==================================%%

\abstract{Air pollution remains a leading global health and environmental risk, particularly in regions vulnerable to episodic air pollution spikes due to wildfires, urban haze and dust storms. 
Accurate forecasting of particulate matter (PM) concentrations is essential to enable timely public health warnings and interventions, yet existing models often underestimate rare but hazardous pollution events. 
Here, we present SynCast, a high-resolution neural forecasting model that integrates meteorological and air composition data to improve predictions of both average and extreme pollution levels. 
Built on a regionally adapted transformer backbone and enhanced with a diffusion-based stochastic refinement module, SynCast captures the nonlinear dynamics driving PM spikes more accurately than existing approaches. 
Leveraging on harmonized ERA5 and CAMS datasets, our model shows substantial gains in forecasting fidelity across multiple PM variables (PM$_1$, PM$_{2.5}$, PM$_{10}$), especially under extreme conditions. 
We demonstrate that conventional loss functions underrepresent distributional tails (rare pollution events) and show that SynCast, guided by domain-aware objectives and extreme value theory, significantly enhances performance in highly impacted regions without compromising global accuracy.  
This approach provides a scalable foundation for next-generation air quality early warning systems and supports climate–health risk mitigation in vulnerable regions. Our code is available at \href{https://github.com/YohanAbeysinghe/Synergistic-Neural-Forecasting-of-Air-Pollution-with-Stochastic-Sampling}{SynCast}.}

%%================================%%
%% Sample for structured abstract %%
%%================================%%


\maketitle

\section{Introduction}\label{sec1}

Air pollution is a leading environmental and public health challenge worldwide, contributing to millions of premature deaths and morbidity each year and causing significant economic and societal disruption~\cite{lelieveld2015contribution, oecd2016economic}. 
Among the pollutants of concern, fine particulate matter (PM\textsubscript{1.0}, PM\textsubscript{2.5}, PM\textsubscript{10.0}) poses a particularly severe threat due to its ability to penetrate into the respiratory system and its strong association with cardiovascular and respiratory diseases~\cite{pope2009, chen2020longterm, yang2020changes}. 
In response to this growing challenge, early warning systems and high-fidelity forecasts of air quality are increasingly considered as vital tools for risk mitigation, public health planning, and environmental governance~\cite{zhang2012realtime}.

Despite the importance of accurate air quality forecasting, existing modeling approaches face persistent limitations. 
Numerical chemistry-transport models such as WRF-Chem \cite{ojha2020widespread}, CMAQ \cite{zhang2012source}, and GEOS-CF \cite{knowland2022nasa} simulate atmospheric pollutant dynamics with high physical fidelity but require intensive computational resources and often exhibit systematic regional biases.
In contrast, recent advances in machine learning (ML) have enabled more computationally efficient models that learn directly from atmospheric reanalysis data~\cite{wang2024, nedungadi2025aircastimprovingairpollution, Aurora2024foundationmodelearth}. 
These data-driven methods offer promising accuracy and scalability and have shown strong performance in capturing average pollution patterns. 
Aurora~\cite{Aurora2024foundationmodelearth} and AirCast~\cite{nedungadi2025aircastimprovingairpollution}, for example, represent important steps toward data-driven air-quality forecasting but remain constrained by deterministic training objectives.
However, a key challenge remains: ML-based models often fail to accurately predict short-lived, high-magnitude pollution events, such as PM spikes caused by wildfires, sandstorms, haze episodes, or elevated anthropogenic emissions~\cite{vela-martin2024large, stirnberg2021meteorology}. 
This limitation is especially problematic in high-impact regions, where public exposure to short-term pollution extremes can cause disproportionate harm.

A key reason for this shortfall is the reliance on symmetric loss functions such as Mean Squared Error (MSE), which emphasize average accuracy while underweighting rare, high-impact deviations. 
As a result, they tend to oversmooth predictions and systematically underestimate extreme values. 
Additionally, the processes driving pollution extremes are governed by complex, nonlinear interactions among meteorology, emissions, and chemistry dynamics.
Traditional deterministic modeling approaches struggle to capture this variability effectively.
Prior efforts to address these challenges have focused on loss function modifications \cite{xu2024extremecastboostingextremevalue, nedungadi2025aircastimprovingairpollution}, stochastic perturbations \cite{fuxiextreme2023}, or ensemble-style predictions. 
However, these methods often fail to capture the skewed, heavy-tailed nature of pollutant distributions, which contrasts with the more symmetric, centrally distributed patterns commonly exhibited by meteorological variables, such as temperature and pressure.

To address these challenges,  we propose \textbf{SynCast}, a hybrid neural forecasting model designed to improve the fidelity of air pollution forecasts under both typical and extreme conditions. 
SynCast builds on a regionally adapted version of the Pangu-Weather \cite{panguweather3dhighresolutionmodel} architecture, extending it to jointly forecast meteorological and air quality variables at high spatiotemporal resolution. 
By incorporating surface and upper-atmospheric air meteorological inputs alongside particulate matter concentrations from reanalysis datasets (ERA5 and CAMS), SynCast captures the coupled dynamics of atmospheric transport and chemical evolution more holistically.

A core innovation of SynCast lies in its generative refinement stage: a diffusion-based module that stochastically enhances deterministic forecasts by generating multiple climatology-aware scenarios. 
Inspired by ensemble-based methods in numerical weather prediction and recent advances in generative modeling~\cite{arches2024, fuxiextreme2023}, this module produces multiple plausible high-resolution scenarios using learned stochastic perturbations and diffusion-based refinement conditioned on climatology and prior forecasts.
This process sharpens the spatial details and better represents uncertainty, particularly in the tails of the distribution.
We further employ parameter-efficient fine-tuning (LoRA \cite{hu2022lora}) to adapt the model for specific regions, enabling flexible adaptation to specific regions of interest without retraining the full network architecture, which incurs substantial computational and data cost.

We evaluate SynCast on regional forecasting tasks with a focus on high-impact events, such as dust storms in the Middle East and haze events in China. 
Results show that SynCast consistently improves accuracy in both average-case and extreme-case scenarios across multiple PM variables, outperforming existing ML and physics-based baselines. 
Our findings demonstrate the value of combining deterministic forecasting with stochastic refinement for localized regional forecasting, and point to the broader potential of hybrid neural models in supporting climate-resilient infrastructure and public health systems in key regions of interest.

\section{Results}\label{sec2}

SynCast delivers significant improvements in both accuracy and robustness for regional air quality forecasting.
For example, across the MENA region, SynCast delivers gains in 24-hour air pollution forecasting, reducing PM$_{2.5}$ RMSE (Root Mean Square Error) by \textbf{$\approx$18--39\%} at both fine and coarse resolutions compared to state-of-the-art approaches Aurora and AirCast (Table~\ref{tab:baseline_comparison}). 
These improvements extend to extreme-event detection: SynCast's diffusion-based enhancement module improves tail-sensitive metrics, reducing PM$_{2.5}$ RQE (Relative Quantile Error) by \textbf{20.8\%} and increasing SEDI (Symmetric Extremal Dependency Index) by \textbf{2.6\%} over its deterministic variant (Table~\ref{tab:ablation}). 
Together, these results underscore the value of combining parameter-efficient regional adaptation with generative refinement to achieve high-fidelity, operational-grade pollution forecasts.

\noindent\textbf{Metrics:}
We use three metrics to quantify performance, with a focus on both general accuracy and extreme event prediction (see Appendix~C for formulations). 

\noindent\textbf{Latitude-Weighted RMSE:} Assesses forecast accuracy across spatial locations, correcting for regional area disparities. We adjust the longitudinal averaging to reflect reduced latitude span, avoiding unfair advantage for local models.

\noindent\textbf{RQE:} Measures deviation at high quantiles (90th–99.99th percentiles), emphasizing the accuracy in capturing extremes. Negative values indicate underestimation.

\noindent\textbf{SEDI:} Captures the skill in classifying extreme vs. non-extreme events, providing robustness across different percentile thresholds.

\begin{table*}[t]
\centering
\footnotesize
\vspace{0.3cm} % Optional vertical space before table
\renewcommand{\arraystretch}{1.2}
\resizebox{0.7\textwidth}{!}{ % Adjust width as needed
\begin{tabular}{l|c|ccc}
\toprule
Model & Resolution & PM$_1$ & PM$_{2.5}$ & PM$_{10}$ \\
\midrule
AirCast     & 5.26° & 6.651  & 8.824  & 13.276 \\
SynCast     & 5.26° & ~\textbf{3.051}  & ~\textbf{5.413}  & ~\textbf{9.583}  \\
\midrule
Aurora      & 0.4°  & 4.324  & 6.781  & 9.285 \\
SynCast     & 0.4°  & ~\textbf{3.201}  & ~\textbf{5.587}  & ~\textbf{8.690}  \\
\bottomrule
\end{tabular}
}
\vspace{0.3cm}
\caption{
Latitude-adjusted RMSE (in $\mu\mathrm{g}/\mathrm{m}^3$) for PM$1$, PM${2.5}$, and PM$_{10}$ across different model configurations over the MENA region, evaluated at their respective native spatial resolutions with a 24-hour lead time. AirCast operates at a coarse resolution of 5.26$^\circ$, while Aurora supports a finer 0.4$^\circ$ resolution. SynCast is evaluated at both resolutions to enable a fair comparison. Results are computed using CAMS forecasts as the reference target. SynCast consistently outperforms both AirCast and Aurora across all PM variables.}
\label{tab:baseline_comparison}
\end{table*}

\begin{figure}[t]
    \centering
    \includegraphics[width=\textwidth]{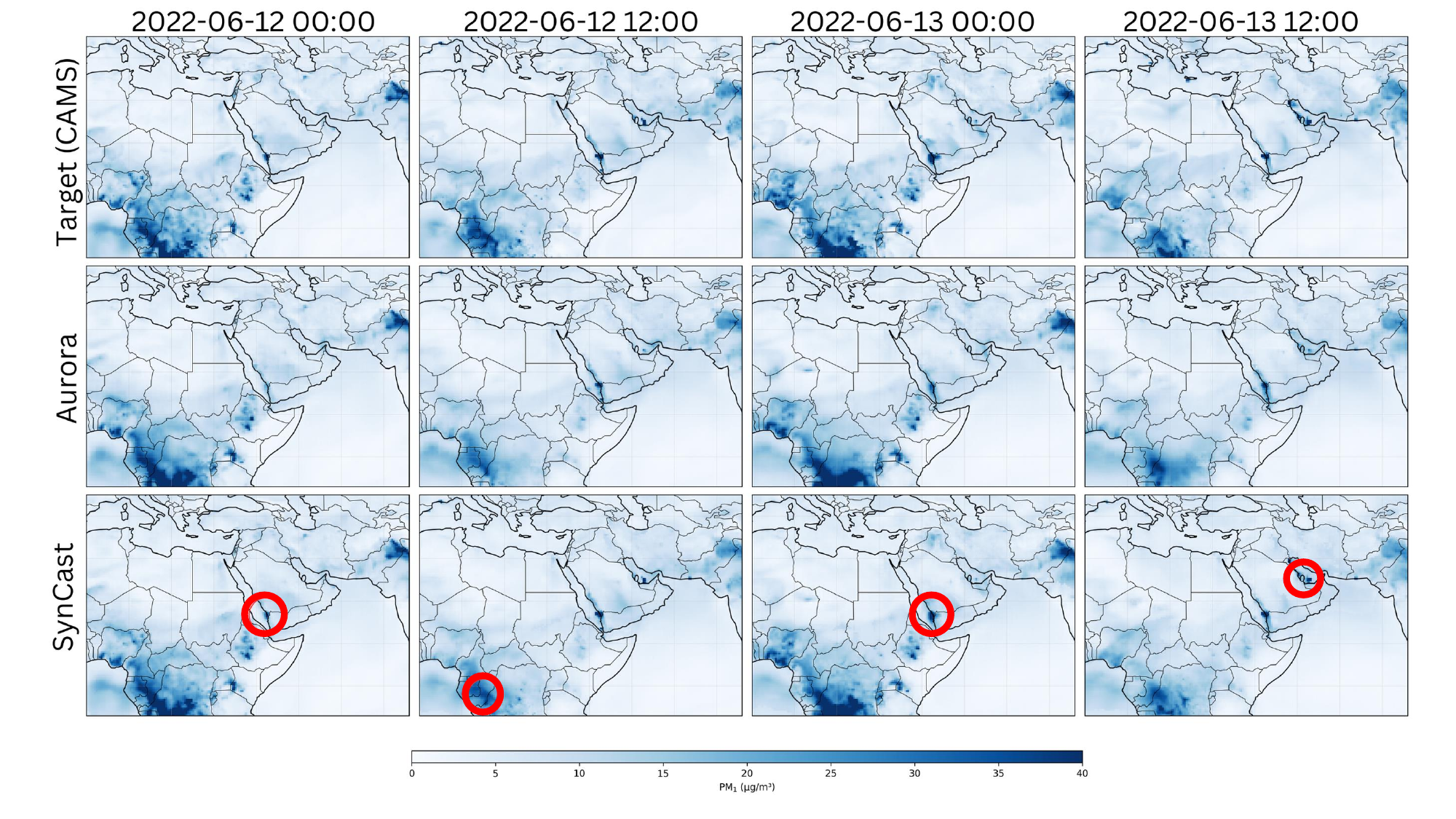}
    \caption{
    We compare PM$_{1}$ forecasts from Aurora and SynCast with CAMS ground truth over four time steps between 12–13 June 2022. While both models capture the overall pollution patterns, SynCast shows better agreement with CAMS, especially in detecting sudden local spikes in pollution levels. These regions, which are missed by Aurora but captured by SynCast, are marked with red circles.
    Additional qualitative comparisons and extended visualizations are provided in Appendix~A}
    \label{fig:baseline_comparison_pm1}
\end{figure}

We benchmark SynCast against two recent deep learning baselines, Aurora~\cite{Aurora2024foundationmodelearth} and AirCast~\cite{nedungadi2025aircastimprovingairpollution}, that support multi-variable PM forecasting.
Aurora produces outputs at $0.4^{\circ}$ resolution, while AirCast operates at a coarser $5.625^{\circ}$ grid. 
To enable a fair comparison, we harmonize SynCast’s predictions by downsampling them using the ClimaX regridding protocol~\cite{nguyen2023climaxfoundationmodelweather}, which ensures spatial alignment across models. 
CAMS forecasts serve as a physics-based operational baseline. 
Table~\ref{tab:baseline_comparison} reports 24-hour performance results, where SynCast consistently achieves the best accuracy across PM$_1$, PM$_{2.5}$, and PM$_{10}$, reflecting the benefit of combining large-scale weather priors with targeted regional adaptation.
Visual comparisons (Figure~\ref{fig:baseline_comparison_pm1}) further show that SynCast preserves sharper pollution gradients and resolves regional high-PM episodes that other models either blur or entirely miss (see Appendix~A for more visualizations). 
Performance on extremes, as measured by RQE and SEDI, highlights SynCast's ability to reduce the underestimation common in deterministic forecasts, with the diffusion-based enhancement contributing most strongly under extreme scenarios (Section~\ref{sec:diff-enhance}).

\begin{figure}[!hbpt]
    \centering
    \includegraphics[width=\textwidth]{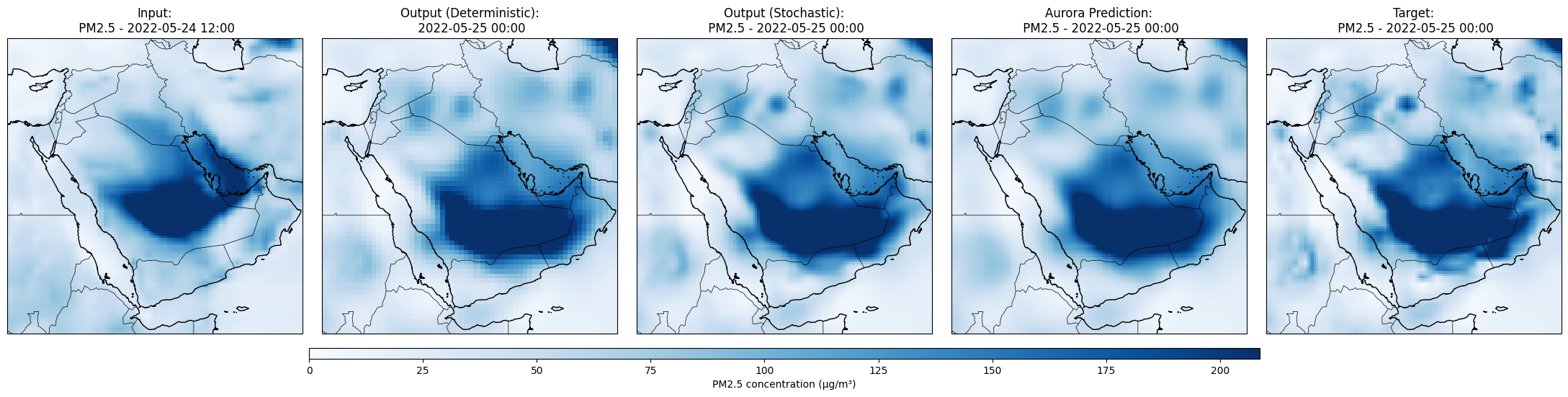}
    \caption{
    $\mathrm{PM}_{2.5}$ predictions during a major dust storm episode in May 2022. The first panel shows input conditions from the previous timestep. The second and third panels present predictions from the deterministic baseline and SynCast with diffusion-based refinement, respectively. The fourth panel shows Aurora forecasts, while the fifth panel displays the CAMS target data. Compared to both the deterministic baseline and Aurora, SynCast more accurately reconstructs the spatial extent and intensity of the dust plume.}
    \label{fig:ExtremeValueFig}
\end{figure}

\noindent\textbf{Extreme Event Predictions: }
To evaluate SynCast's ability to capture high-impact pollution episodes, we examine a representative PM\textsubscript{2.5} event over the Middle East during late May 2022, previously documented in the literature~\cite{FRANCIS2023119539}.
The region experienced dust storm that mobilized huge amounts of dust from Iraq and Syria and transported it downstream across Kuwait, Saudi Arabia, and the UAE. 
The event resulted in record aerosol loads, hazardous visibility reductions, and significant radiative impacts, making it one of the most severe multi-day dust outbreaks in recent decades.
We conduct inference using both the deterministic backbone and SynCast (with its diffusion-based enhancement) on this event. As shown in Figure~\ref{fig:ExtremeValueFig}, the deterministic model underestimates peak intensities and produces spatially smoother predictions. In contrast, SynCast reconstructs sharper gradients and more coherent plume structures, aligning better with the reference distribution. These results highlight the benefit of the stochastic refinement module in resolving sharp transitions and rare pollution spikes.

\begin{figure}[t]
    \centering
    \includegraphics[width=0.8\textwidth]{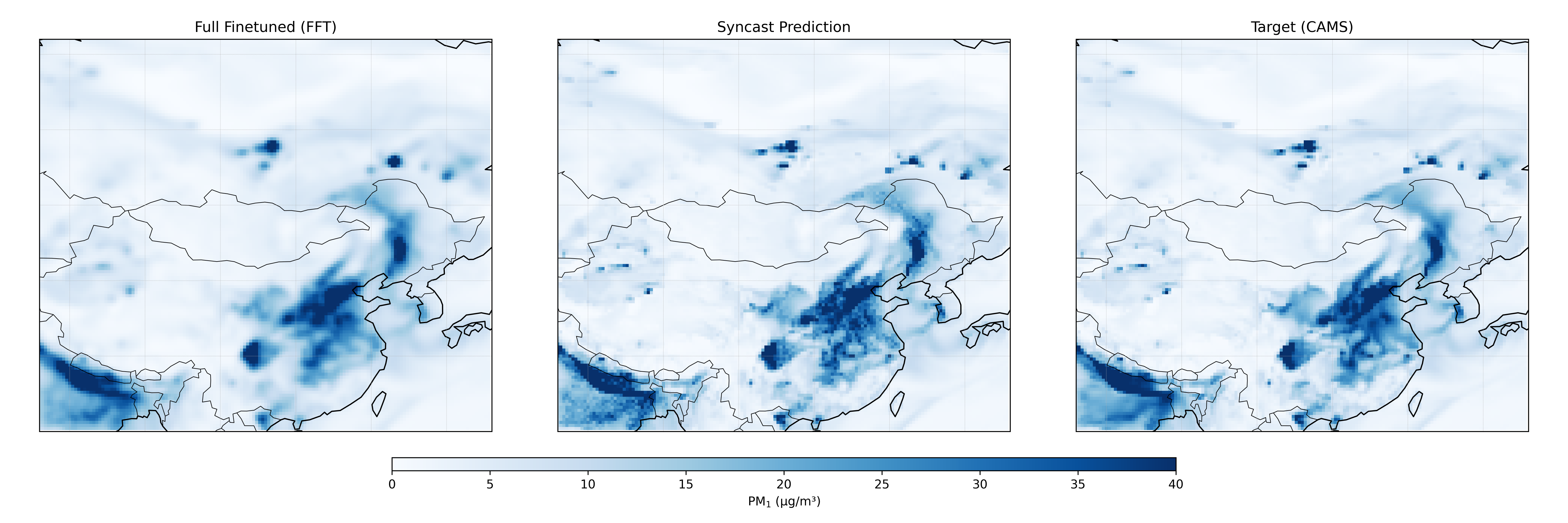}
    \caption{Generalization results over the Chinese region. SynCast was trained exclusively on the MENA domain but is evaluated here for PM$_1$ forecasts in China. The first panel shows outputs from a globally full fine-tuned model (FFT), which tends to oversmooth local structures. In contrast, SynCast (middle) better preserves sharp gradients and plume coherence, capturing broad spatial patterns and major pollution events with closer alignment to the CAMS target (right). 
    Some localized degradation remains, but the results highlight SynCast's stronger transferability relative to naive global fine-tuning.}
    \label{fig:generalization_results}
\end{figure}

\noindent\textbf{Generalization Results: }
While SynCast is fine-tuned on the MENA region, we evaluate its generalization capacity by running inference on geographically distinct regions not seen during training. This test assesses the retention of the pre-trained backbone's global knowledge and the extent of extrapolation beyond the fine-tuned domain.
Table~\ref{tab:generalization_results} summarizes the results, showing that SynCast maintains competitive accuracy in unseen regions, which suggests effective transfer from the foundation model. However, a modest performance drop is observed compared to the MENA domain, underscoring the value of region-specific adaptation. Interestingly, regions with similar seasonal cycles and meteorological drivers of PM extremes, such as China (Figure~\ref{fig:generalization_results}), exhibit stronger generalization, indicating that similarity in atmospheric dynamics facilitates transferability.

\begin{table*}[t]
\centering
\footnotesize
\vspace{0.3cm} % Optional vertical space before table
\renewcommand{\arraystretch}{1.2}
\resizebox{0.75\textwidth}{!}{ % Adjust width as needed
\begin{tabular}{l|l|ccc}
\toprule
Region & Model & PM$_1$ & PM$_{2.5}$ & PM$_{10}$ \\
\midrule
\multirow{2}{*}{Chinese Region} 
    & Full finetuned (FFT) & 4.923 & 7.700 & 12.250 \\
    & SynCast & 4.851 & 7.623 & 12.128 \\
\midrule
\multirow{2}{*}{European Region} 
    & Full finetuned (FFT) & 5.120 & 9.100 & 14.450 \\
    & SynCast & 5.058 & 9.050 & 14.329 \\
\bottomrule
\end{tabular}
}
\vspace{0.3cm}
\caption{Latitude-weighted RMSE for PM$_1$, PM$_{2.5}$, and PM$_{10}$ in regions excluded from training. The SynCast model was trained on the MENA region and evaluated on the Chinese and European regions to assess generalization. Results for two models are shown for each region in comparison to full finetuned global model.}
\label{tab:generalization_results}
\end{table*}

\noindent\textbf{Impact of Model Components:} To understand the effect of each architectural and training component, we conducted targeted experiments by modifying or omitting individual design elements. Removing regional cropping or incorporating new variables via full fine-tuning caused substantial performance drops, primarily due to catastrophic forgetting \cite{kemker2018measuring} of the original meteorological features (Table~\ref{tab:ablation_main}, Rows 2 and 3). In contrast, parameter-efficient adaptation through LoRA~\cite{hu2022lora} preserved prior knowledge while enabling accurate PM predictions (Row 5). Excluding meteorological inputs led to markedly higher errors, as shown in Row 4, underscoring their importance for guiding PM estimation. Furthermore, using PM variables alone with LoRA (Row 6) underperformed compared to setups that included meteorological features, probably due to the pretrained Pangu~\cite{panguweather3dhighresolutionmodel} backbone's reliance on these inputs.
Although the RMSE gains from the diffusion enhancement were nominal (Table~\ref{tab:ablation_diffusion}), significant improvements were observed in metrics sensitive to extremes, including RQE~\cite{pathak2022fourcastnetglobaldatadrivenhighresolution} and SEDI~\cite{ferro2011extremal} (Table~\ref{tab:ablation}). This supports the effectiveness of the diffusion stage in refining predictions for rare, high-impact pollution events.

\begin{table*}[t]
\centering
\vspace{0.3cm}
\renewcommand{\arraystretch}{1.2}
\resizebox{1.0\textwidth}{!}{
\begin{tabular}{l|cccc|ccccccc}
\toprule
\textbf{Model Configuration} & \textbf{PM Vars} & \textbf{Surface Vars} & \textbf{LoRA} & \textbf{Regional Adaptation} & 
\textbf{mslp} & \textbf{u10} & \textbf{v10} & \textbf{t2m} & \textbf{PM$_1$} & \textbf{PM$_{2.5}$} & \textbf{PM$_{10}$} \\
\midrule
\textbf{Baseline (Pangu)} & \xmark & \cmark & \xmark & \xmark & 59.071 & 1.742 & 1.843 & 0.633 & - & - & - \\
\textbf{Full Finetune (FFT)} & \cmark & \cmark & \xmark & \xmark & 4078.276 & 5.684 & 4.859 & 23.160 & 9.076 & 14.254 & 18.388 \\
\textbf{Regional FFT} & \cmark & \cmark & \xmark & \cmark & 4003.711 & 5.769 & 4.925 & 22.812 & 9.222 & 14.028 & 18.721 \\
\textbf{PM-Only Regional FFT} & \cmark & \xmark & \xmark & \cmark & - & - & - & - & 9.127 & 14.087 & 18.674 \\
\textbf{LoRA FT} & \cmark & \cmark & \cmark & \xmark & 2705.325 & 8.289 & 5.603 & 24.363 & 3.966 & 6.772 & 11.858 \\
\textbf{PM-Only Regional LoRA FT} & \cmark & \xmark & \cmark & \cmark & - & - & - & - & 5.552 & 9.481 & 16.601 \\
\textbf{SynCast (Ours, w/o DEE)} & \cmark & \cmark & \cmark & \cmark & 59.366 & 1.336 & 1.218 & 0.563 & 3.288 & 5.656 & 8.772 \\
\bottomrule
\end{tabular}
}
\vspace{0.3cm}
\caption{Performance of different adaptation strategies on the MENA region with a 24-hour lead time, using ERA5 and CAMS as reference datasets. 
The first four columns indicate whether the model includes PM variables, meteorological surface variables, parameter-efficient fine-tuning with LoRA (Low-Rank Adaptation), or regional adaptation through cropping. 
Model configurations are as follows: 
Baseline, 
\text{FFT} (Full Fine-Tuning); 
\text{Regional FFT}; 
\text{PM-Only Regional FFT}; 
\text{LoRA FT}; and 
\text{PM-Only Regional LoRA FT}. 
The final row (\text{SynCast w/o DEE}) represents our configuration without the diffusion-based extreme enhancement (DEE) module. 
The remaining columns report RMSE scores ($\mu$g/m$^3$) for meteorological variables (mean sea level pressure [mslp], wind at 10 m [u10], wind at 10 m [v10], and temperature at 2 m [t2m]) as well as for PMs (PM$_1$, PM$_{2.5}$, and PM$_{10}$).}
\label{tab:ablation_main}
\end{table*}

\begin{table*}[t]
\centering
\vspace{0.3cm}
\renewcommand{\arraystretch}{1.2}
\resizebox{0.85\textwidth}{!}{ % Adjust width as needed
\begin{tabular}{l|cccc|ccc}
\toprule
\textbf{Model Configuration} & \textbf{PM Vars} & \textbf{Surface Vars} & \textbf{LoRA} & \textbf{Diffusion (DEE)} & 
\textbf{PM$_1$} & \textbf{PM$_{2.5}$} & \textbf{PM$_{10}$} \\
\midrule
\textbf{SynCast (Ours, w/o DEE)} & \cmark & \cmark & \cmark & \xmark & 3.288 & 5.656 & 8.772 \\
\textbf{SynCast (Ours, with DEE)} & \cmark & \cmark & \cmark & \cmark & \textbf{3.201} & \textbf{5.587} & \textbf{8.690} \\
\bottomrule
\end{tabular}
}
\vspace{0.3cm}
\caption{Impact of the diffusion-based extreme enhancement (DEE) module. All models are trained on the MENA region and evaluated on 24-hour lead forecasts using ERA5 and CAMS as targets. RMSE is reported in $\mu$g/m$^3$ for all PM variables.}
\label{tab:ablation_diffusion}
\end{table*}

\begin{table*}[t]
\centering
% \footnotesize
\vspace{0.3cm} % Optional vertical space before table
\renewcommand{\arraystretch}{1.2}
\resizebox{0.85\textwidth}{!}{ % Reduced width since fewer columns
\begin{tabular}{l|ccc|ccc}
\toprule
\multirow{2}{*}{Model Configuration} & \multicolumn{3}{c|}{RQE} & \multicolumn{3}{c}{SEDI} \\
\cmidrule(lr){2-4} \cmidrule(lr){5-7}
 & PM$_1$ & PM$_{2.5}$ & PM$_{10}$ & PM$_1$@90th & PM$_{2.5}$@90th & PM$_{10}$@90th \\
\midrule
\textbf{SynCast (Ours, w/o DEE)} & -0.0023 & -0.0024 & -0.0029 & 0.6841 & 0.6975 & 0.6889 \\
\textbf{SynCast (Ours, with DEE)} & -0.0017 & -0.0019 & -0.0020 & 0.7032 & 0.7158 & 0.7075 \\
\bottomrule
\end{tabular}
}
\vspace{0.3cm} % Optional vertical space after table before caption
\caption{Results evaluating SynCast with and without the diffusion-based extreme enhancement (DEE) module. RQE (Relative Quantile Error) measures the accuracy of predicted pollutant magnitudes, where negative values indicate underestimation. SEDI (Symmetric Extremal Dependence Index) assesses the model's ability to detect extreme events (90th percentile and above), with values closer to 1 indicating better detection. Results are reported across PM$_1$, PM$_{2.5}$, and PM$_{10}$.
Experiment showing RQE (- underestimate, + overestimate) and SEDI (the closer to 1, the better) scores on PM variables.
}
\label{tab:ablation}
\end{table*}

\begin{figure}[t]
    \centering
    \includegraphics[width=\textwidth]{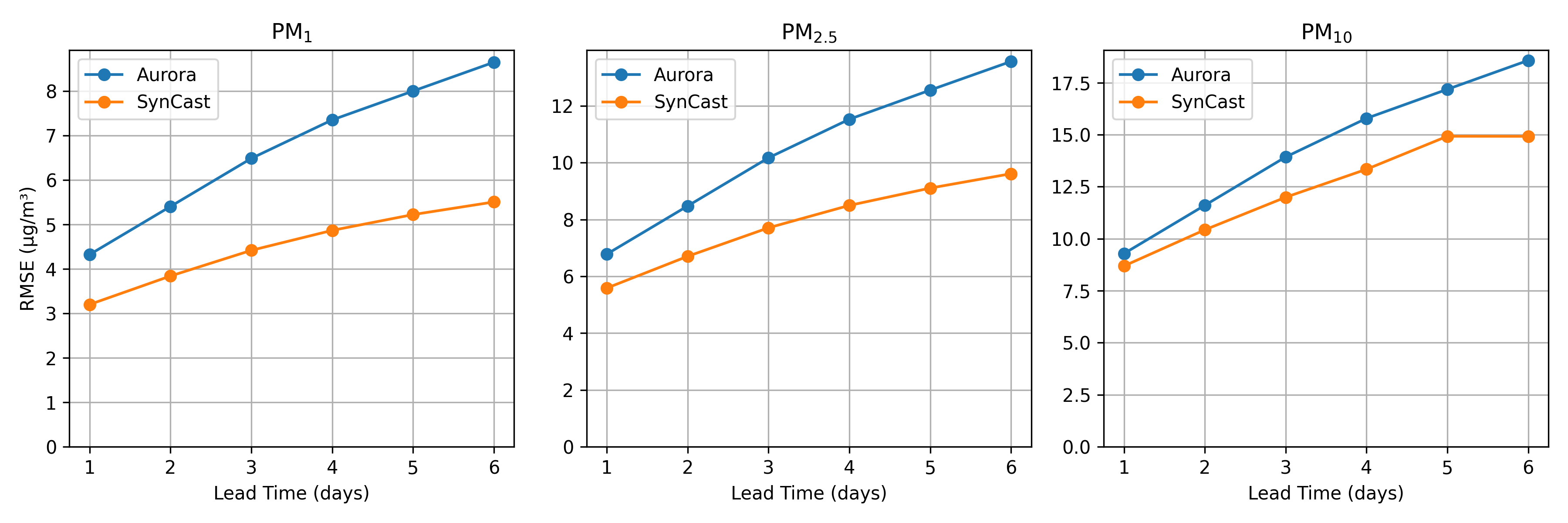}
    \caption{
    Performance comparison of SynCast and Aurora across increasing lead times (1-6 days) on PM$_1$, PM$_{2.5}$, and PM$_{10}$ forecasting over the MENA region. RMSE ($\mu$g/m$^3$) is reported using CAMS as ground truth. SynCast consistently outperforms Aurora across all time horizons and particulate matter types, demonstrating more stable and accurate long-range predictions.
    }
    \label{fig:rollout_rmse}
\end{figure}

\noindent\textbf{Forecasting Over Multi-Day Lead Times: }
To further assess SynCast’s robustness across varying temporal horizons, we analyze its performance over extended lead times ranging from 1 to 6 days. As shown in Figure~\ref{fig:rollout_rmse}, SynCast consistently achieves lower RMSE values than Aurora across all PM variables, PM$_1$, PM$_{2.5}$, and PM$_{10}$, demonstrating superior performance in long-range forecasting. 
The error margin between the two models widens with increasing lead time, underscoring SynCast’s ability to retain accuracy in more challenging forecasting horizons. 
This pattern suggests that the model is not merely learning statistical associations but also leveraging physically consistent relationships, enabling more robust generalization to large-scale dynamics.
This trend reinforces the utility of our design choices, such as parameter-efficient tuning and regional specialization, for stable and reliable air quality modeling in operational contexts.

\begin{figure}[t]
    \centering
    \includegraphics[width=\textwidth]{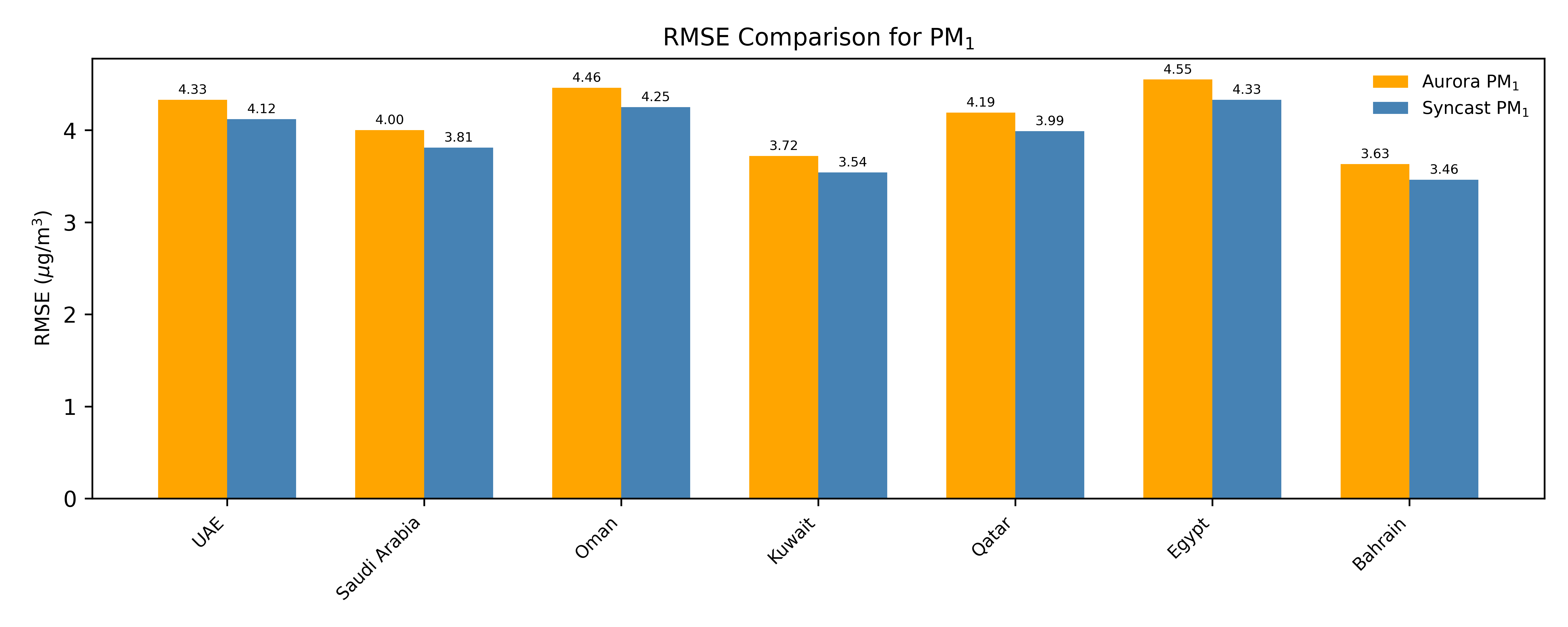}
    \caption{RMSE comparison for PM$_1$ predictions across several Middle Eastern countries, evaluating SynCast against Aurora. SynCast consistently achieves lower RMSE, demonstrating its improved generalization and regional adaptability in country-level forecasts.}
    \label{fig:country_wise_rmse}
\end{figure}

\noindent\textbf{Country-wise Performance:} 
We further assess the generalization capability of SynCast at the country level within the MENA region.
As shown in Figure~\ref{fig:country_wise_rmse}, SynCast consistently achieves lower RMSE values for PM$_1$ across evaluated countries, including the UAE, Saudi Arabia, Oman, and Egypt. 
These gains indicate that the model not only captures broad regional dynamics but also adapts effectively to localized emission sources and meteorological conditions specific to each country.
The ability to provide accurate forecasts at national granularity is particularly important for practical deployment. 
Improved fidelity at this scale enhances the operational utility of early warning systems, supports targeted mitigation strategies, and strengthens the responsiveness of public health and environmental agencies.
This country-level robustness underscores SynCast's suitability for downstream use in heterogeneous settings where policy decisions and alerts are issued at national or sub-national levels.

\section{Discussion}

\begin{figure}[t]
    \centering
    \includegraphics[width=0.8\textwidth]{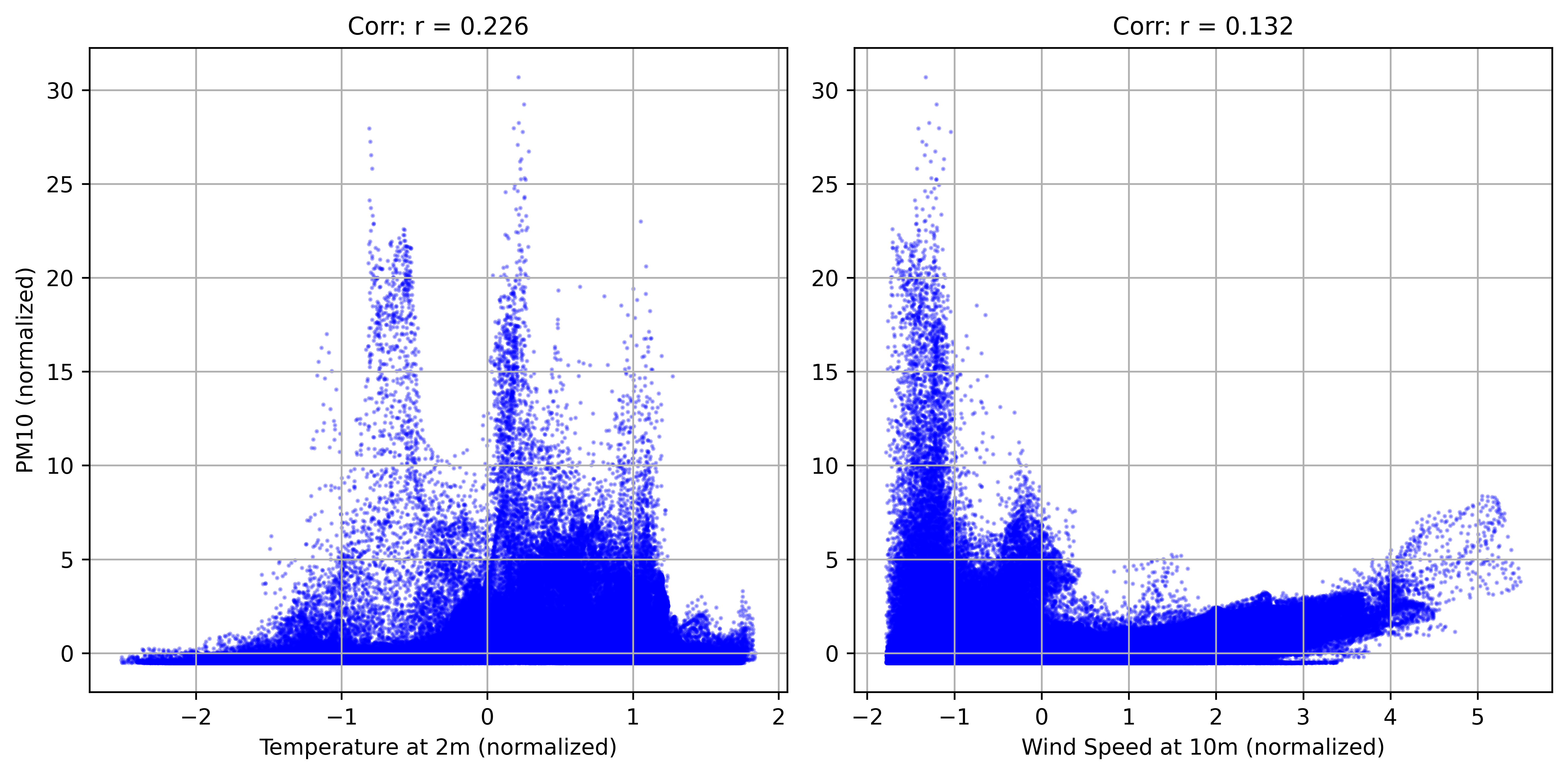}
    \caption{
    Correlation scatter plots between PM$_{10}$ values and selected meteorological surface variables, temperature at 2m (left) and wind speed at 10m (right). Although the individual correlations appear moderate, they are statistically significant and reflect meaningful interactions, such as pollutant dispersion and aerosol formation.
    }
    \label{fig:correlation}
\end{figure}

Our findings demonstrate that SynCast offers notable improvements in both general and extreme PM forecasting across diverse spatial domains. 
The integration of meteorological variables, especially temperature, wind components, and pressure, proves essential to stabilize learning and improve predictive accuracy. 
This is consistent with both our empirical observations and prior studies~\cite{gidarjati2024correlation}, confirming that PM concentrations are tightly linked to dynamic atmospheric conditions, albeit through moderate and often nonlinear correlations. 
Specifically, Figure~\ref{fig:correlation} shows that temperature and wind speed exhibit Pearson correlation coefficients with PM$_{10}$, respectively. 
Even though these correlations are only moderate, they show that weather still plays an important role in PM levels. 
Higher temperatures can make the air more stable and limit vertical mixing, which traps pollutants near the surface. 
Wind speed, on the other hand, controls how much pollutants spread out: strong winds help disperse them, while calm winds allow them to build up. Together, these factors determine whether pollution episodes intensify or clear out.
Deep learning models can capture nonlinear interactions among meteorological drivers, such as how temperature, wind, and humidity jointly shape dispersion and accumulation, which linear correlation measures cannot fully reflect.

\begin{figure}[t]
    \centering
    \includegraphics[width=0.8\textwidth]{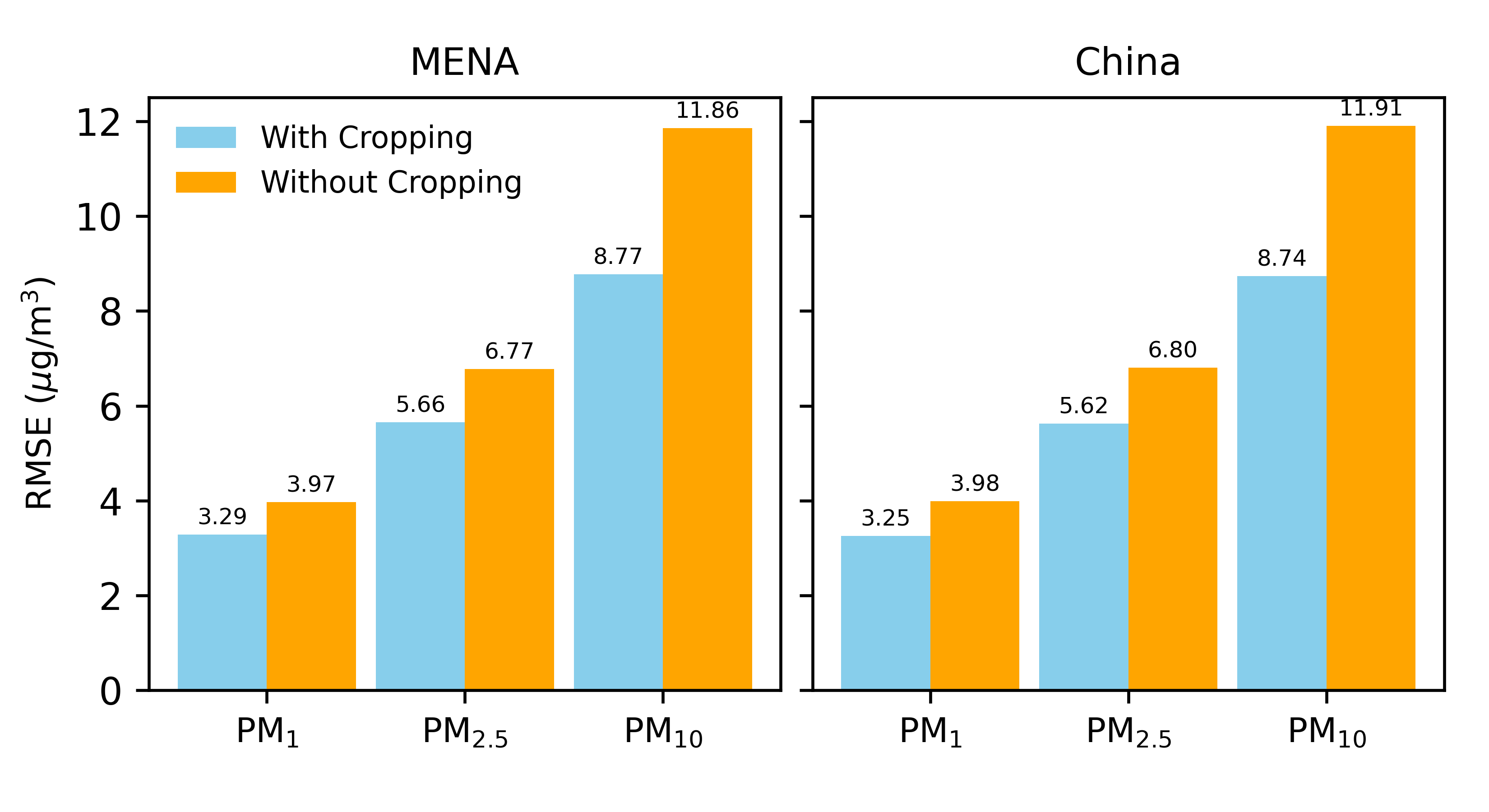}
    \caption{
    Impact of region-specific training on PM forecasting over the MENA and China regions. Models are trained using regionally cropped data (blue) or without cropping (orange), following Table~\ref{tab:ablation_main} settings. RMSE ($\mu$g/m$^3$) is reported for PM$_1$, PM$_{2.5}$, and PM$_{10}$. Region-specific training consistently improves accuracy across all pollutant types, underscoring the benefit of tailoring model learning to local meteorological and emission characteristics.
    }
    \label{fig:localregion_cropping_barchart}
\end{figure}

Crucially, our component-wise evaluation (Table~\ref{tab:ablation_main}) reveals that performance degrades significantly when full fine-tuning is applied naively, leading to catastrophic forgetting \cite{kemker2018measuring}. These trends reinforce the importance of preserving pretrained representations and utilizing parameter-efficient fine-tuning strategies such as LoRA~\cite{hu2022lora}.

Moreover, regional cropping emerges as a simple yet effective mechanism to boost local fidelity in predictions. As shown in Figure~\ref{fig:localregion_cropping_barchart}, tailoring the training process to specific geographic domains significantly enhances representational precision and improves convergence. This is evident from the reduction in RMSE across all PM types when cropping is applied during training for both the MENA and China regions. These results follow the configuration outlined in Table~\ref{tab:ablation_main}. The improvements highlight the benefit of allowing the model to specialize in local emission patterns, topographical influences, and meteorological dynamics. 
SynCast’s ability to adapt to specific regions through focused training not only improves accuracy but can also be applied to other areas with different weather and pollution patterns.

The diffusion-based enhancement module provides only marginal gains in RMSE, but significantly improves tail-sensitive metrics like RQE and SEDI (Table~\ref{tab:ablation}). These results underscore its utility in capturing sharp gradients and localized peaks, properties that are critical for anticipating high-impact air quality events. As demonstrated in the May 2022 Middle East dust storm (Figure~\ref{fig:ExtremeValueFig}), SynCast more accurately resolves pollutant plumes and intensity gradients, suggesting improved preparedness for public health and environmental interventions.

Taken together, these findings affirm the effectiveness of SynCast’s modular and region-aware architecture. 
By combining a transformer-based backbone with parameter-efficient tuning and a conditional diffusion stage, SynCast not only advances average forecasting accuracy but also substantially improves sensitivity to extremes. 
The framework leverages large-scale weather priors while remaining adaptable to regional domains, positioning it as a scalable solution for next-generation air quality forecasting and environmental risk assessment.

\noindent\textbf{Limitations \& Future Directions: }
A key limitation of our study is the reliance on reanalysis products (ERA5 and CAMS) as the reference ground truth. While these datasets provide globally consistent coverage, they are not direct observations and may propagate systematic biases into forecasts, particularly for aerosol loads. 
Moreover, ERA5 and CAMS are not available in real time, which constrains their applicability for operational or near-real-time early-warning systems. Although the diffusion enhancement in SynCast is selectively triggered to reduce overhead, it still adds computational cost that may challenge strict operational timelines. 
Several directions are promising. First, bias correction pipelines and integration of ground and satellite observations should be incorporated to mitigate dependence on reanalysis alone. 
Second, SynCast could be extended to forecast additional atmospheric variables and air pollutants, enabling a more holistic environmental risk framework. 
Finally, integrating meteorological, chemical, and observational streams in a multimodal design could strengthen SynCast’s generalization and support globally scalable, locally precise forecasting.

\begin{figure}[t]
    \centering
    \includegraphics[width=\textwidth]{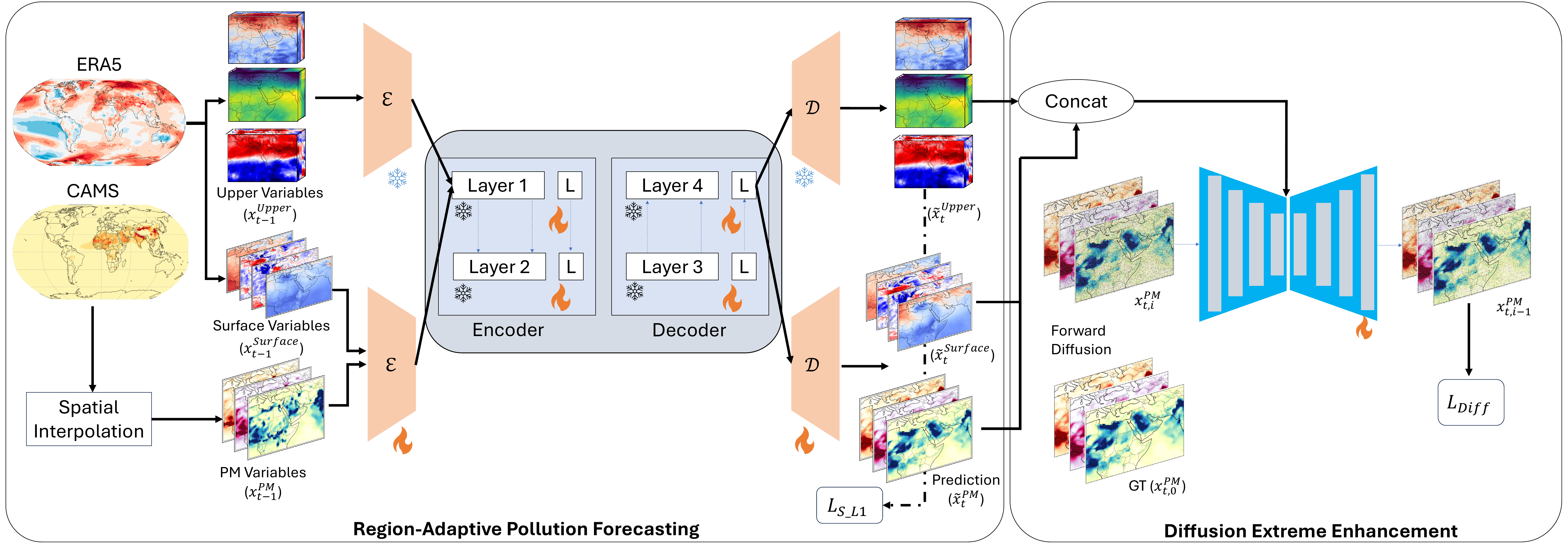}
    \caption{Overview of the SynCast architecture. The system ingests meteorological variables from ERA5 (surface and upper-air) and particulate matter concentrations from CAMS, spatially interpolated and processed into unified inputs. 
    To enhance high-impact event prediction, a diffusion module refines these outputs by reconstructing sharp, localized structures in extreme pollution distributions. 
    The flame symbol indicates loss functions applied during training, including for deterministic predictions and diffusion loss for the refinement stage.
    The Region-Adaptive Pollution Forecasting model is trained using the Smooth L1 loss, which allows it to capture general patterns from common values while handling the outliers.
    }
    \label{fig:main}
\end{figure}

\section{Method}\label{secmethod}

SynCast combines ERA5 meteorological inputs and CAMS air quality fields in a transformer-based encoder–decoder for spatiotemporal PM forecasting (Figure~\ref{fig:main}). 
Unlike existing foundation models such as Pangu-Weather~\cite{panguweather3dhighresolutionmodel}, Fuxi~\cite{sun2024fuxi}, FourCastNet~\cite{pathak2022fourcastnetglobaldatadrivenhighresolution}, and GraphCast~\cite{graphcastlearningskillfulmediumrange}, which are restricted to weather variables, SynCast directly incorporates CAMS-derived particulate matter concentrations, enabling joint modeling of weather-pollution dynamics. 
Furthermore, in contrast to recent PM-focused baselines like Aurora~\cite{Aurora2024foundationmodelearth} and AirCast~\cite{nedungadi2025aircastimprovingairpollution}, SynCast introduces a diffusion-based refinement stage that improves sharpness and accuracy under extreme pollution events. 
These design choices collectively address the failure modes of prior models, which struggle with extremes due to deterministic training.

\subsection*{Problem Formulation and Notation} 
We formulate air pollution forecasting as a \textit{conditional spatiotemporal prediction task}, focusing specifically on particulate matter (PM) variables; PM$_1$, PM$_{2.5}$ and PM$_{10}$, which are critical to public health, yet are often underrepresented in traditional forecast workflows. 
The objective is to predict future surface-level concentrations of these PM species at a target time step $t$, conditioned on the preceding atmospheric state.

At each time step \( t \), the atmospheric input comprises two components: upper-air variables and surface-level variables. The upper-air variables are represented as a 4D tensor \( \mathbf{X}_t^{\text{upper}} \in \mathbb{R}^{N \times Z \times H \times W} \), where \( N \) denotes the batch size, \( Z \) is the number of pressure levels, and \( H, W \) are the spatial dimensions (latitude and longitude, respectively). The surface-level variables are denoted as \( \mathbf{X}_t^{\text{surface}} \in \mathbb{R}^{N \times H \times W} \), as they are defined only at the earth's surface.
We denote the full atmospheric state as \( \mathbf{X}_t = \{\mathbf{X}_t^{\text{upper}}, \mathbf{X}_t^{\text{surface}}\} \), and define a predictive model \( F_\theta \), parameterized by weights \( \theta \), which learns the conditional mapping \(
F_\theta(\mathbf{X}_t) = P(\mathbf{X}_t \mid \mathbf{X}_{t-1}).
\)
This naturally extends to autoregressive multi-step forecasting, where the model is recursively applied to forecast further into the future, expressed as \( \mathbf{X}_{t+n} = F_\theta^{(n)}(\mathbf{X}_t) \), with \( n \in \mathbb{Z}^+ \).
We first form the extended surface-level input as $\mathbf{X}_t^{\text{surface}} = \{ \mathbf{X}_t^{\text{surface}}, \mathbf{X}_t^{\text{PM}} \}$, and subsequently extract sub-domains of size $(H_l, W_l)$ at the native 0.25$^\circ$ resolution for region-specific forecasting, enabling joint learning of meteorological and pollution dynamics within spatially coherent patches.

\noindent\textbf{Data and Setup:} Modern forecasting systems rely on high-resolution reanalysis products. We use ERA5~\cite{haiden2021evaluation} as the primary meteorological input, which provides hourly global fields at $0.25^{\circ} \times 0.25^{\circ}$ resolution, forming a regular grid of size $H=721, W=1440$. To incorporate air quality dynamics, we augment ERA5 with pollution estimates from the ECMWF Atmospheric Composition Reanalysis 4 (EAC4), part of the Copernicus Atmosphere Monitoring Service (CAMS)~\cite{CAMS23}.
EAC4 is natively available at $0.75^{\circ}$ resolution and 3-hourly intervals. We harmonize it with ERA5 by applying bilinear spatial interpolation to $0.25^{\circ}$ and temporal upsampling to hourly frequency. This alignment enables coherent learning over meteorological and air composition variables. For PM-related inputs, we denote the aligned particulate matter fields as $\mathbf{X}^{\text{PM}}_t \in \mathbb{R}^{N \times H \times W \times 3}$ at each time step $i$, consistent with the spatial grid used for $\mathbf{X}_t^{\text{surface}}$.

Training spans 2003-2019, where 2003-2017 is used for training, 2020–2021 for validation, and 2019 for testing, consistent with prior protocols~\cite{panguweather3dhighresolutionmodel}. The model uses 12 total input variables: 5 upper-air variables across 13 pressure levels and 7 surface-level variables, including meteorological (t2m, u10, v10, msl) and air quality indicators (PM$_1$, PM$_{2.5}$, PM$_{10}$). Table~\ref{tab:variables} summarizes these variables and their sources.
For region-specific fine-tuning, we extract cropped patches of size $(H_l, W_l)$ while preserving native spatial resolution. This facilitates localized adaptation in high-impact areas such as East Asia and MENA. Additionally, we apply a diffusion-based generative refinement module focused on capturing extreme events in PM variables.

\begin{table*}[t]
\centering
\begin{tabular}{llll}
\toprule
\textbf{Variable} & \textbf{Description} & \textbf{Levels} & \textbf{Source} \\
\midrule
\rowcolor{gray!15}
\multicolumn{4}{l}{\textit{Surface (Single-Level) Variables}} \\
u10        & Zonal wind at 10 m height               & Surface     & ERA5 \\
v10        & Meridional wind at 10 m height          & Surface     & ERA5 \\
t2m        & Air temperature at 2 m height           & Surface     & ERA5 \\
msl        & Mean sea level pressure                 & Surface     & ERA5 \\
PM$_1$     & Particulate matter $<$1 $\mu$m          & Surface     & CAMS \\
PM$_{2.5}$ & Particulate matter $<$2.5 $\mu$m        & Surface     & CAMS \\
PM$_{10}$  & Particulate matter $<$10 $\mu$m         & Surface     & CAMS \\
\midrule
\rowcolor{gray!15}
\multicolumn{4}{l}{\textit{Upper-Air (Multi-Level) Variables (13 pressure levels)}} \\
z          & Geopotential height                     & 13 levels   & ERA5 \\
q          & Specific humidity                       & 13 levels   & ERA5 \\
u          & Zonal wind                              & 13 levels   & ERA5 \\
v          & Meridional wind                         & 13 levels   & ERA5 \\
t          & Air temperature                         & 13 levels   & ERA5 \\
\bottomrule
\end{tabular}
\caption{List of meteorological and air composition variables used in SynCast. PM variables are from CAMS; all others are from ERA5.}
\label{tab:variables}
\end{table*}

\subsection*{Region-Adaptive Pollution Forecasting}
To extend high-resolution weather forecasting to include air quality variables and enhance performance over specific regions, we build on the 3D Earth-specific Transformer (3DEST) backbone from Pangu-Weather~\cite{panguweather3dhighresolutionmodel}. Our model introduces architectural adaptations to support PM forecasting and efficient regional fine-tuning.
For a lead time $\Delta t$ (e.g., 24 hours), the model processes two input tensors: upper-level variables across 13 pressure levels and surface-level variables that include both meteorological and PM fields. The inputs are defined over a localized MENA domain of size $(H_l, W_l) = (217, 312)$, corresponding to latitudes $[-7^\circ, 45^\circ]$ and longitudes $[0^\circ, 76^\circ]$. The upper-level tensor is shaped as $13 \times W_l \times H_l \times 5$, and the extended surface input is $W_l \times H_l \times 7$.

\noindent\textbf{Patch Embedding and Decoder:}
To enable transformer-based modeling, we apply patch embedding. Upper-level inputs are split using a $2 \times 4 \times 4$ patch size, producing an embedded tensor of shape $7 \times W_e \times H_e \times C$, while surface-level inputs use a $4 \times 4$ patch size, yielding $W_e \times H_e \times C$. These are concatenated to form a unified tensor of shape $8 \times W_e \times H_e \times C$ (with $W_e = 55$, $H_e = 78$), passed through a transformer encoder-decoder, and projected back to the original spatial resolution via patch recovery.

\noindent\textbf{Patch Recovery:} Patch recovery performs the inverse operation of patch embedding, but does not share its parameters. This process reduces the matrix depth while restoring the spatial resolution to the original input dimensions.

\noindent\textbf{Regional Positional Bias:}
We retain spatial inductive biases through the Earth-specific positional bias matrix $B$ used in the attention mechanism. For regional modeling, we crop $B$ to match the $(H_l, W_l)$ grid, preserving localized geospatial structure.

\noindent\textbf{Parameter-Efficient Fine-Tuning:}
To adapt the pretrained global model to new variables and regional domains, we apply LoRA~\cite{hu2022lora}, a parameter-efficient fine-tuning (PEFT) technique. LoRA introduces low-rank adapters \( \mathbf{A}, \mathbf{B} \) into frozen linear layers. With rank \( r = 8 \), scaling factor \( \alpha = 16 \), and dropout rate of 0.1, the updated projection becomes:
\begin{equation}
\mathbf{W}_\text{base} + \Delta \mathbf{W} = \mathbf{W}_\text{base} + \mathbf{B}\mathbf{A},
\end{equation}
where \( \mathbf{W}_\text{base} \in \mathbb{R}^{d \times k} \) is a frozen pretrained matrix, and \( \mathbf{A} \in \mathbb{R}^{r \times k}, \mathbf{B} \in \mathbb{R}^{d \times r} \) are learnable low-rank matrices.

\noindent\textbf{Log Transformation for PM Variables:}
PM concentrations exhibit heavy-tailed distributions, where extreme values can dominate optimization and hinder stable training. 
To mitigate this, we apply a log-based transformation to compress high-magnitude outliers and enhance variability in low-concentration regimes:
\begin{equation}
x_{\text{log}} = \frac{\log(\max(x, 10^{-11})) - \log(10^{-11})}{\log(10^{-4})},
\end{equation}
where $x$ denotes the raw PM concentration. 
This normalization maps values to the $[0, 1]$ range, limits the effect of extreme spikes, and improves learning for typical concentration ranges.

\noindent\textbf{Loss Function: } Let \( x_{\text{log}} \) and \( \hat{x}_{\text{log}} \) denote the log-transformed ground truth and predicted PM values, respectively. 
We optimize SynCast using the Smooth L1 loss \cite{ren2015faster}:
\begin{equation}
\mathcal{L}_{\text{SmoothL1}} = \frac{1}{N} \sum_{i=1}^{N} \text{SmoothL1}\left(x_{\text{log}}^{(i)}, \hat{x}_{\text{log}}^{(i)}\right),
\end{equation}
where \( N \) is the number of spatial grid points (i.e., pixels in the regridded domain). 
This loss behaves like L2 for small errors, encouraging stability, and like L1 for large errors, ensuring robustness to outliers. 
Together with the log transformation, it prevents the model from being dominated by noisy spikes. 
We emphasize, however, that tail sensitivity is not addressed at this stage; improvements in forecasting extremes are achieved primarily through the subsequent diffusion-based enhancement module.

\subsection*{Diffusion-based Extreme Enhancement}
\label{sec:diff-enhance}
While the region-adaptive pollution forecasting module performs well under typical air quality conditions, its accuracy tends to degrade during \textit{extreme pollution events}, a known limitation in high-impact scenario forecasting~\cite{fuxiextreme2023,xu2024extremecastboostingextremevalue}. 
To address this, we introduce a diffusion-based extreme enhancement stage that refines the region-adaptive predicted PM outputs. 
This stage leverages a denoising diffusion model to correct underestimation in rare, high-magnitude PM regimes, using the region-adaptive outputs as context. 
Importantly, diffusion refinement is selectively triggered only under extreme-event settings, ensuring computational efficiency and making SynCast practical for operational early-warning applications.

\noindent\textbf{Modeling Setup.}
Let the diffusion model be parameterized by \( \Theta \). We define a noisy PM sample at diffusion timestep \( i \in \{1, \dots, N\} \) as \( \mathbf{x}_{t,i}^{\text{PM}} \), generated from the ground-truth field \( \mathbf{x}_{t,0}^{\text{PM}} \) via a predefined noise schedule. The model is trained to reconstruct the added noise \( \boldsymbol{\epsilon} \) using the standard denoising score-matching loss:
\begin{equation}
    \mathcal{L}_{\text{Diff}} = \mathbb{E}_{i, \boldsymbol{\epsilon}, \mathbf{x}, \mathbf{c}} \left\| \epsilon_{\Theta}(\mathbf{x}_{t,i}^{\text{PM}}, \mathbf{c}) - \boldsymbol{\epsilon} \right\|^2,
\end{equation}
where the conditioning vector \( \mathbf{c} \in \mathbb{R}^{W_l \times H_l \times C} \) consists of the outputs of the region-adaptive pollution forecasting:
\[
\mathbf{c} = \left[ \mathbf{X}_t^{\text{PM}}, \mathbf{X}_t^{\text{surface}}, \mathbf{X}_t^{\text{upper}} \right],
\]
which include the model-predicted PM field, surface-level variables, and upper-air variables. These serve as spatial priors for guiding refinement in challenging regimes.

\noindent\textbf{Inference.}
At test time, the diffusion model refines \( \mathbf{X}_t^{\text{PM}} \) through iterative denoising starting from a stochastic initialization. The refined output replaces the initial region-adaptive predicted PM field, while predictions for all other atmospheric variables remain unchanged.

\noindent\textbf{Climatology-Aware Sampling.}
To avoid unnecessary refinement under normal conditions, the diffusion enhancement is triggered selectively based on deviations from region-specific climatology. 
Specifically, if the predicted PM concentration exceeds its regional climatological threshold, the diffusion model is activated:
\begin{equation}
\text{if } \mathbf{X}_t^{\text{PM}} > \mathbf{X}_t^{\text{Climatology}} + \delta \quad \text{then} \quad \hat{\mathbf{X}}_t^{\text{PM}} = \mathbf{X}_{t,0}^{\text{PM}}; \quad \text{else} \quad \hat{\mathbf{X}}_t^{\text{PM}} = \mathbf{X}_t^{\text{PM}}.
\end{equation}
Here, \( \delta \) is a tunable threshold and \( \hat{\mathbf{X}}_t^{\text{PM}} \) denotes the final PM prediction. This pixel level hybrid strategy ensures computational efficiency and stability while improving fidelity in rare, high-pollution scenarios.

\newpage
\noindent\textbf{Data Availability: }
ERA5 reanalysis data are publicly available from the Copernicus Climate Data Store (\url{https://cds.climate.copernicus.eu/}). CAMS atmospheric composition data are available from the Copernicus Atmosphere Monitoring Service (\url{https://ads.atmosphere.copernicus.eu/}). Processed datasets used in this study are available from the corresponding author upon request. 

\noindent\textbf{Code Availability: }
The SynCast code will be provided by the corresponding author upon request.

\backmatter

% \bmhead{Acknowledgements}
\noindent\textbf{Acknowledgements: }
This research was partially supported by the MBZUAI-WIS Joint Program for Artificial Intelligence Research.

\noindent\textbf{Funding: }
This research was partially supported by the MBZUAI-WIS Joint Program for Artificial Intelligence Research.

\noindent\textbf{Author contributions: }
Y. A. developed the methodology, conducted the experiments, aligned the data, prepared visualizations, and contributed to the manuscript writing. M. A. M. contributed to methodology development, writing, analysis, and manuscript revisions. S. B. supported the formulation of the diffusion module and helped in preparing visualizations. R. S., F. S. K., and Y. R. provided critical feedback and revisions. S. K. supervised the project, guided the research direction, and contributed to writing, feedback, and revisions. All authors reviewed and approved the final manuscript.

\noindent\textbf{Competing interests: }
The authors declare that they have no competing interests.

\noindent\textbf{Supplementary information}

\noindent\textbf{Additional Information: }
Available in the Appendix sections (Visualizations in Appendix~A, Implementation Details in Appendix~B and Formulations in Appendix~C).

\noindent\textbf{Correspondence} and requests for materials should be addressed to Muhammad Akhtar Munir.

\newpage
\bibliography{sn-bibliography}% common bib file
%% if required, the content of .bbl file can be included here once bbl is generated
%%\input sn-article.bbl

\newpage
\begin{appendices}
% \noindent\textbf{Section title of first appendix}
\section{More qualitative results}\label{secA1}

\begin{figure}[h]
    \centering
    \includegraphics[width=\textwidth]{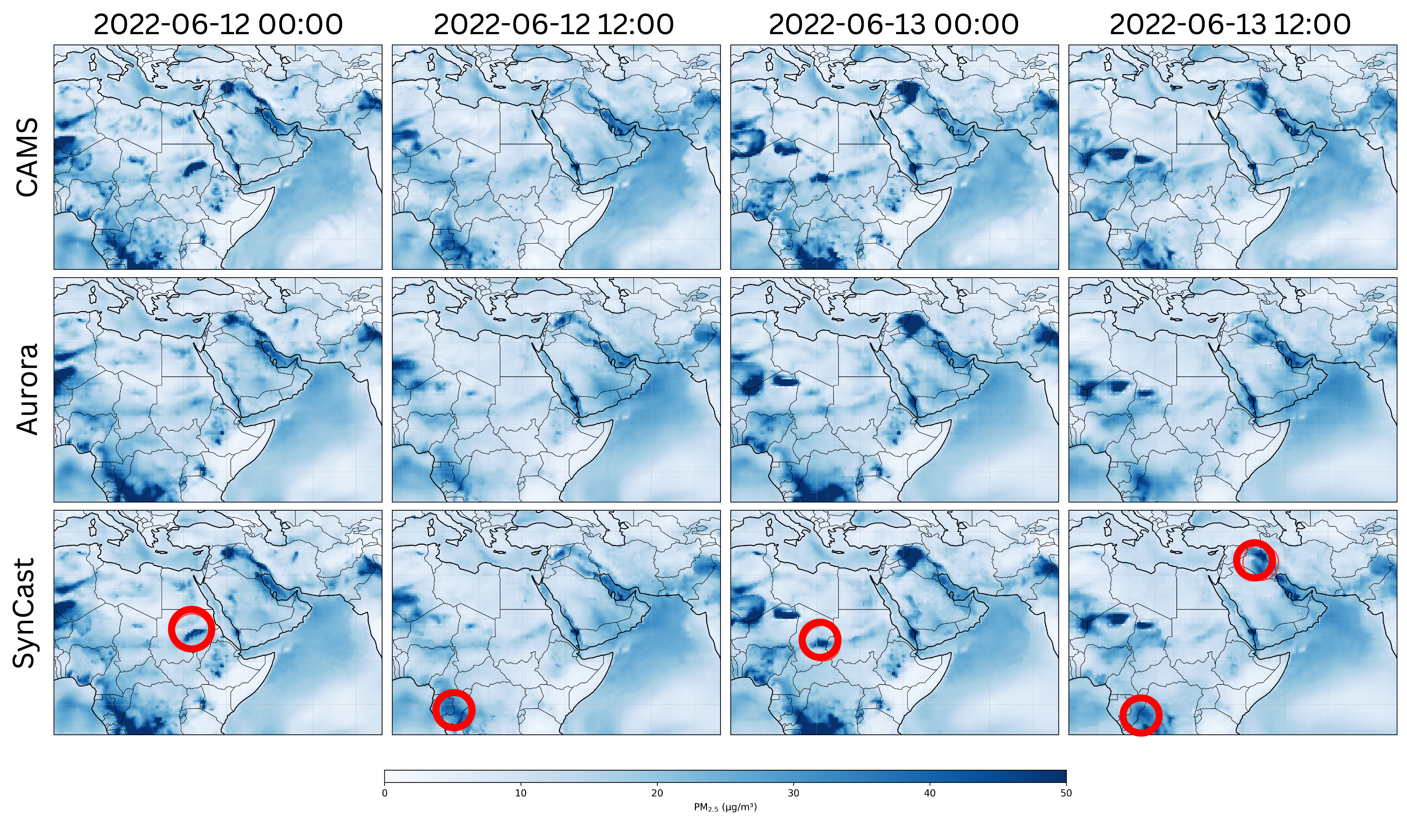}
    \caption{Baseline comparison of PM$_{2.5}$ forecasts from Aurora and SynCast against CAMS ground truth over four time steps (12–13 June 2022). 
    While both models capture the broad spatial patterns of pollution, SynCast better resolves localized high-PM episodes (circled in red) that Aurora fails to detect.}
    \label{fig:baseline_comparison_pm2p5}
\end{figure}

\begin{figure}[h]
    \centering
    \includegraphics[width=\textwidth]{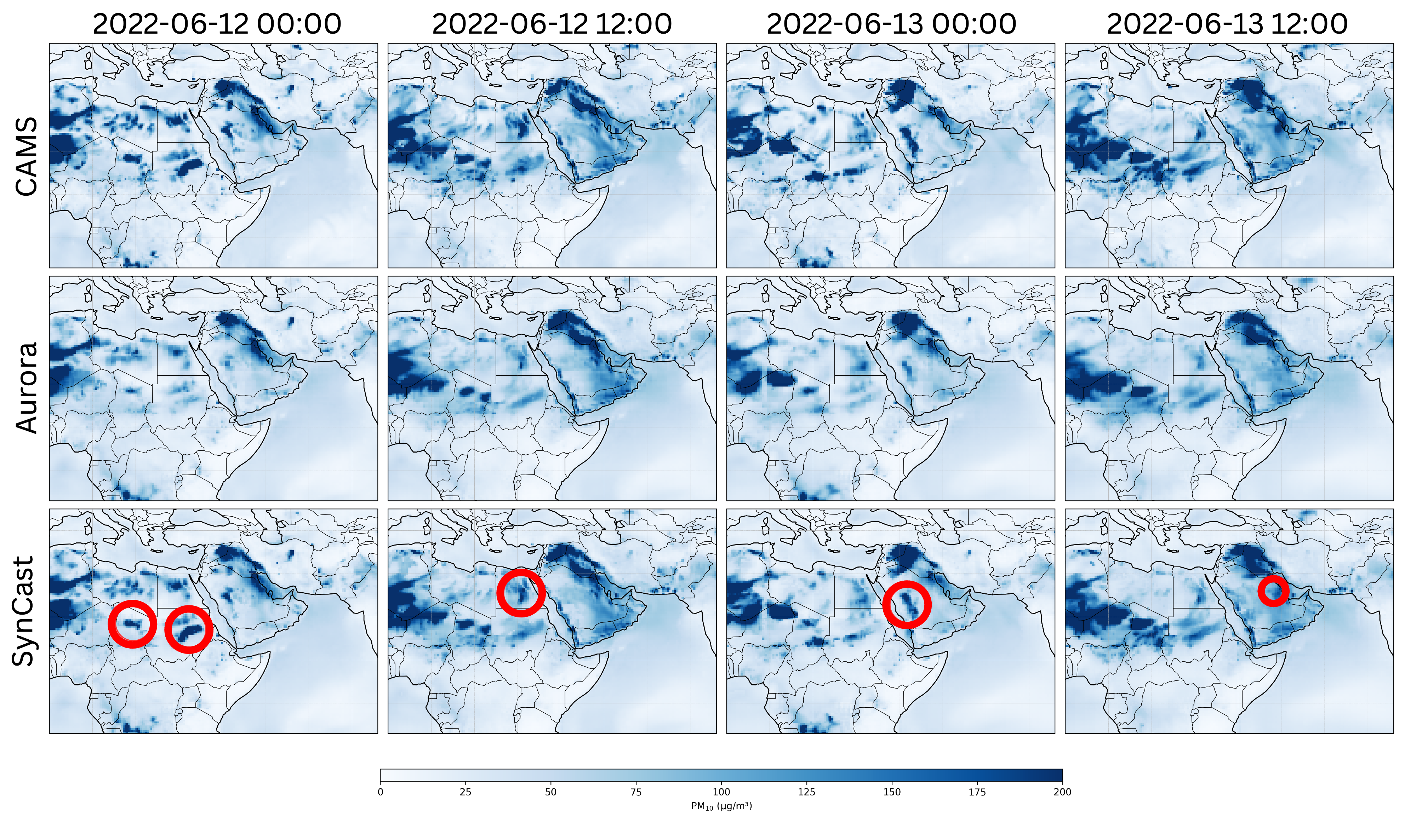}
    \caption{Baseline comparison of PM$_{10}$ forecasts from Aurora and SynCast against CAMS ground truth over four time steps (12–13 June 2022). Similar to PM$_{2.5}$, SynCast more accurately reproduces localized dust plumes and high-PM events (circled in red), which Aurora underestimates or smooths out.}
    \label{fig:baseline_comparison_pm10}
\end{figure}

We extend the baseline evaluation to PM$_{2.5}$ and PM$_{10}$ variables, as shown in Figures~\ref{fig:baseline_comparison_pm2p5} and \ref{fig:baseline_comparison_pm10}. 
While Aurora captures the broad distribution of pollution, it tends to underestimate localized high-intensity events. 
In contrast, SynCast shows improved agreement with CAMS ground truth, particularly in reproducing sharp, localized pollution spikes. 
These improvements are especially evident in PM$_{10}$, where SynCast resolves dust plume intensities and regional hotspots that Aurora diffuses or misses entirely. 
Together, these results highlight SynCast’s superior ability to capture both fine-scale and large-scale particulate matter dynamics across pollutant types.

\section{Implementation Details}\label{secA2}

The deterministic component of our module is trained on four NVIDIA A100 GPUs (40 GB each). After applying the final cropped version of our model, \textbf{SynCast (Ours, w/o DEE)}, one epoch requires approximately 1--2 hours. A key advantage of cropping is the improved memory efficiency: whereas previously only a single batch could be accommodated on each GPU, the current setup allows processing of up to 8 batches per GPU. The learning rate is set to $1 \times 10^{-5}$. For the loss function, we assign different weights to the surface variables [\textit{msl}, \textit{u10}, \textit{v10}, \textit{t2m}, \textit{pm1}, \textit{pm25}, \textit{pm10}] in the ratio [1.50, 0.77, 0.66, 3.00, 1.20, 1.20, 1.20]. The training is initialized with Pangu-pretrained weights and is fine-tuned for a total of 20 epochs, requiring approximately one day in total.

The diffusion-based component also benefits from cropping. Without cropping, it was not possible to fit the entire global map at 0.25$^\circ$ resolution into a 40~GB NVIDIA A100 GPU. With the current cropped setting, the model is fine-tuned for 5 epochs using a learning rate of $1 \times 10^{-6}$ and a batch size of 1.

\section{Formulas}\label{secA3}

\subsection*{Relative Quartile Error (RQE)}

The \textbf{Relative Quartile Error (RQE)} measures the relative deviation of the predicted quartiles from the observed quartiles. It is defined as:

\[
\text{RQE} = \frac{1}{3} \sum_{i=1}^{3} \frac{Q_i^{\text{pred}} - Q_i^{\text{obs}}}{Q_i^{\text{obs}}}
\]

where \(Q_i^{\text{obs}}\) and \(Q_i^{\text{pred}}\) are the $i$-th quartiles of the observed and predicted data, respectively, with $i = 1,2,3$ corresponding to the 25th percentile, median, and 75th percentile.

\subsection*{Symmetric Extremal Dependence Index (SEDI)}

The \textbf{Symmetric Extremal Dependence Index (SEDI)} evaluates the skill of a model in predicting extreme events. It is computed from a contingency table and focuses on hits and false alarms. SEDI is symmetric, ranges from $-1$ to $1$, and is suitable for rare events.

\[
\text{SEDI} = \frac{\ln(F) - \ln(H) - \ln(1-F) + \ln(1-H)}{\ln(F) + \ln(H) + \ln(1-F) + \ln(1-H)}
\]

where
\[
H = \frac{a}{a+c} \quad \text{(hit rate)}, \qquad
F = \frac{b}{b+d} \quad \text{(false alarm rate)}
\]

Here, $a$, $b$, $c$, and $d$ are elements of the contingency table:

\begin{itemize}
  \item $a$ = hits (predicted and observed extremes)
  \item $b$ = false alarms (predicted extreme but not observed)
  \item $c$ = misses (observed extreme but not predicted)
  \item $d$ = correct negatives (neither predicted nor observed extreme)
\end{itemize}

\subsection*{Latitude-Weighted Root Mean Square Error (Lat-Weighted RMSE)}

The \textbf{Latitude-Weighted Root Mean Square Error (RMSE)} evaluates model errors on global gridded data while accounting for the varying area represented by each latitude. Since the surface area of grid cells decreases toward the poles, weighting by the cosine of latitude ensures that errors near the poles do not disproportionately influence the overall RMSE.

\[
\text{Lat-Weighted RMSE} = \sqrt{ \frac{\sum_{i,j} w_i \, (P_{i,j} - O_{i,j})^2}{\sum_{i,j} w_i} }
\]

where:
\begin{itemize}
  \item $P_{i,j}$ = predicted value at latitude $i$, longitude $j$
  \item $O_{i,j}$ = observed value at latitude $i$, longitude $j$
  \item $w_i = \cos(\phi_i)$ = latitude weight based on the latitude $\phi_i$ in radians
\end{itemize}

\subsection*{Smooth L1 Loss}

The \textbf{Smooth L1 Loss} (also known as Huber Loss) is a loss function that combines the advantages of L1 and L2 losses. It behaves like an L2 loss for small errors (to ensure smooth gradients) and like an L1 loss for large errors (to reduce sensitivity to outliers). This makes it particularly suitable for regression tasks where outliers may exist.

\[
\text{Smooth L1 Loss}(x, y) =
\begin{cases}
0.5 \, (x - y)^2 & \text{if } |x - y| < \delta,\\[2mm]
\delta \, (|x - y| - 0.5 \, \delta) & \text{otherwise}
\end{cases}
\]

where:
\begin{itemize}
  \item $x$ = predicted value
  \item $y$ = target (observed) value
  \item $\delta$ = transition point between L1 and L2 behavior
\end{itemize}

\end{appendices}

\end{document}